%% file: main.tex
\pdfoutput=1
\documentclass[11pt]{article}
\usepackage[preprint]{acl}
\usepackage{times}
\usepackage{latexsym}
\usepackage[T1]{fontenc}
\usepackage[utf8]{inputenc}
\usepackage{microtype}
\usepackage{inconsolata}
\usepackage{graphicx}
\usepackage{booktabs}
\usepackage{amsmath}
\usepackage{amssymb}
\usepackage{multirow}
\usepackage{tikz}
\usetikzlibrary{arrows.meta,positioning,fit,backgrounds,calc}
\usepackage{xcolor}

\title{Label Smoothing in Disguise?\\
Gated Knowledge Distillation for Low-Resource Bangla Generation}

\author{Dipto Sumit, Zobeida Zaman Chowdhury, Syeda Fabiayee Jahan, Farig Sadeque \\
  BRAC University, Dhaka, Bangladesh \\
  \texttt{dipto.sumit@g.bracu.ac.bd}}

\begin{document}
\maketitle

\begin{abstract}
Gated knowledge distillation (KD) weights teacher signals per token or example, aiming to trust
an imperfect teacher only where it seems reliable. We test two gates: EWAD (entropy-weighted
agreement-aware distillation), which gates tokens by teacher confidence, and CHAD
(counterfactual harm-aware distillation), which gates examples by gradient alignment with
validation data. Across six Bangla-centric tasks, three student models (60M to about 2B
parameters), and two teacher families, no gate significantly improves plain KD in 45
comparisons. KD also fails to beat cross-entropy (CE) fine-tuning on gold outputs in 40
single-seed comparisons or three-seed runs on four tasks; CE leads on every summarisation
metric. The main benefit of KD is regularisation: over ten fixed epochs, CE students lose 0.030
to 0.044 token-F1 from their early-stopped checkpoint, versus 0.005 to 0.025 for distilled
students. Teacher-free label smoothing closes most of this gap. In one comparison, a teacher
0.29 token-F1 worse yields students at most 0.009 worse. Thus, gating may change regularisation
strength but has little teacher knowledge to select. Findings are limited to Bangla-centric
tasks and small students. We will release code, data splits, predictions, and scores.\end{abstract}

\section{Introduction}
\label{sec:intro}

Knowledge distillation (KD) trains a small \emph{student} model to match the output
distribution of a larger \emph{teacher} \citep{hinton2015distilling}.
\citet{kim-rush-2016-sequence} extended it to sequence generation, and it is now a standard way
to compress language models
\citep{sanh2020distilbertdistilledversionbert,jiao-etal-2020-tinybert,wang2020minilm}. The
premise is that a distilled student outperforms the same student fine-tuned with cross-entropy
(CE) on the gold outputs alone.

For low-resource languages, this premise is harder to rely on \citep{joshi-etal-2020-state}.
Teachers are trained on little data and make more mistakes, and the capacity gap between
teacher and student is often large. One response is to \emph{gate} the KD signal, trusting the
teacher more where it seems reliable \citep{wen-etal-2023-f,koo-etal-2025-switch}. For Bangla
summarisation, \citet{sumit2026when} proposed two such gates. EWAD (entropy-weighted
agreement-aware distillation)\footnote{We keep the original names. Because each name covers
several objectives in the original paper and code, we name the variants we test by what they
gate on (EWAD-conf, EWAD-agree).} lowers the weight of the KD loss on tokens where the teacher
is not confident or where two teachers disagree. CHAD (counterfactual harm-aware distillation)
lowers the weight of training examples whose KD gradient points away from the gradient of the
validation loss, which was the case for roughly half of the BanSum training data. Both gates
were reported to improve summarisation, with the main results coming from single runs on one
task with one student.

We study these two gates for three reasons. They are, to our knowledge, the only gated KD
methods designed for and evaluated on low-resource generation. They cover the two main kinds of
gate: per token, on teacher confidence or agreement, and per example, on an estimate of training
influence. And their positive results make them a fair test of whether gating helps. We do not
test other ways of selecting the teacher signal, such as choosing among divergences
\citep{wen-etal-2023-f} or switching between teacher and student \citep{koo-etal-2025-switch}.

We evaluate both gates on six Bangla-centric tasks with three students and two teacher
families, and we measure seed and implementation variation explicitly. The reported gains do
not carry over to this setting. In 45 comparisons, no gate significantly beats plain KD
(\S\ref{sec:parity}), and the differences are the size of seed noise (\S\ref{sec:floors}).
Distillation of any kind also fails to beat CE fine-tuning of the same student
(\S\ref{sec:ce}). We argue that KD acts here mainly as a regulariser. It slows overfitting when
the student is trained for many epochs, and label smoothing without a teacher reproduces most of
that effect (\S\ref{sec:regularise}). The quality of the teacher barely reaches the student
(\S\ref{sec:transfer}). A gate applied to such a signal can change the amount of
regularisation, but it has little teacher knowledge to select. Summarisation, the task the
gates were designed for, shows the same pattern (\S\ref{sec:summ}), and
\S\ref{sec:replication} discusses how our results relate to the original ones. All tasks are
Bangla-centric, the students are small, and the teachers are fine-tuned on the same data as the
students. Our claims are restricted to this setting.

\section{Related Work}
\label{sec:related}

\paragraph{Distillation and gating.} Beyond output-level KD
\citep{hinton2015distilling,kim-rush-2016-sequence}, students can be distilled from
intermediate layers \citep{jiao-etal-2020-tinybert,sanh2020distilbertdistilledversionbert},
from relations between examples \citep{Park_2019_CVPR}, from several teachers
\citep{10.1145/3097983.3098135,fukuda17_interspeech} or through a teacher assistant
\citep{mirzadeh2019improvedknowledgedistillationteacher}, including for summarisation
\citep{shleifer2020pretrainedsummarizationdistillation}. These methods weight every example
and token equally. Selective methods choose among divergences \citep{wen-etal-2023-f}, switch
between teacher and student guidance by token-level disagreement \citep{koo-etal-2025-switch},
or gate by teacher confidence and agreement (EWAD) or by gradient alignment with validation
data (CHAD; \citealp{sumit2026when}). CHAD's score is an estimate of training-data influence
\citep{pmlr-v70-koh17a,NEURIPS2020_e6385d39}. All of these methods assume that the KD signal
carries example- or token-specific knowledge worth selecting. This is the assumption we test.

\paragraph{KD for low-resource languages.} Distillation is widely used to build smaller models
for low-resource languages, for example in machine translation together with transfer learning
\citep{dabre-fujita-2020-combining}, in multilingual sentence encoders and bitext mining
\citep{reimers-gurevych-2020-making,heffernan-etal-2022-bitext}, and in compact
language-specific models for cross-lingual transfer \citep{ansell-etal-2023-distilling}. Our
question is narrower: whether, for a pretrained student fine-tuned on a few thousand examples,
the teacher signal adds anything beyond the gold labels, and whether gating it helps.

\paragraph{KD as regularisation.} Several studies argue that much of KD's benefit is
regularisation. Label smoothing and KD have related effects on representations
\citep{muller2019label}, and confidence penalties act as output regularisers
\citep{pereyra2017regularizing}. A poorly trained teacher, or even a hand-designed smooth
distribution, gives most of KD's gain \citep{yuan2020revisiting}. Self-distillation behaves like
instance-specific label smoothing \citep{zhang2020selfdistillation,tang2020understanding}, and
students often do not match their teachers closely even when they generalise well
\citep{stanton2021does}. These studies are mostly on image classification. We test the question
for low-resource generation, with a weak-teacher comparison and a direct label-smoothing
control.

\paragraph{Variance.} Reported NLP gains often fall within the variation caused by seeds, data
order or implementation choices
\citep{reimers-gurevych-2017-reporting,dodge-etal-2019-show,dodge2020finetuning,bouthillier2021accounting}.
We measure these sources for our own pipeline and compare every effect against them.

\section{Methods Under Test}
\label{sec:methods}

\paragraph{Notation.} Let $x$ be an input with gold output $y=(y_1,\dots,y_L)$. At position
$t$ the student $\theta$ predicts $p^S_\theta(\cdot\mid y_{<t},x)$ and the teacher
$p^T(\cdot\mid y_{<t},x)$. The CE and KD losses at $t$ are
\begin{align}
\ell^{\text{CE}}_t &= -\log p^S_\theta(y_t\mid y_{<t},x),\\
\ell^{\text{KD}}_t &= \tau^2\,\mathrm{KL}\!\left(p^T_\tau \,\big\|\, p^S_{\theta,\tau}\right),
\end{align}
with temperature $\tau$. The BanglaT5 student uses its teacher's full distribution. The two
large language model (LLM) students (\S\ref{sec:models}) read the teacher's top-20
log-probabilities from a cache; the KL is computed exactly
over those tokens, and the remaining mass is treated as one tail bucket. All methods share the
student, data, optimiser, schedule and decoding. They differ only in how the KD loss is weighted
(Table~\ref{tab:methods}).

\begin{table}[t]
\centering\small
\setlength{\tabcolsep}{4pt}
\resizebox{\columnwidth}{!}{%
\begin{tabular}{@{}lll@{}}
\toprule
method & weight on the KD loss & used in \\
\midrule
CE & none & all students \\
plain KD & constant $\lambda$ & all students \\
EWAD-conf & per token, teacher confidence & 5 tasks \\
EWAD-agree & per token, teacher agreement & summ.\ \\
\;+\,CPDP & adds a second-teacher term & all students \\
CHAD & per example, gradient alignment & 5 tasks \\
entropy, semantic gate & per example, heuristic & BanglaT5 \\
agreement-harm gate & per token, agreement & summ.\ \\
\bottomrule
\end{tabular}}
\caption{Methods compared. EWAD, CPDP (capacity-proportional divergence preservation) and CHAD
follow \citet{sumit2026when}. The entropy and semantic gates are cheaper variants of CHAD's
gate, and the agreement-harm gate is a forward-only proxy of CHAD used for summarisation.
Details and diagrams are in Appendix~\ref{app:methods}.}
\label{tab:methods}
\end{table}

\paragraph{CE and plain KD.} CE trains on $\mathcal{L}_{\text{CE}}=\sum_t\ell^{\text{CE}}_t$
and never sees a teacher. Plain KD adds the KD loss with weight $\lambda=0.5$ at $\tau=2$,
$\mathcal{L}_{\text{KD}}=\sum_t(\ell^{\text{CE}}_t+\lambda\,\ell^{\text{KD}}_t)$. Because of
the factor $\tau^2$, the KD term carries twice the weight of the CE term.

\paragraph{EWAD-conf.} This gate trusts the teacher only where it is confident. With the
teacher's top-1 probability $p^T_{\max,t}$,
\begin{align}
g_t &= \sigma\big(k\,(p^T_{\max,t}-\delta)\big), \quad \alpha_t = g_t\,(1-\eta),\\
\mathcal{L}_{\text{EWAD}} &= \textstyle\sum_t \big((1-\alpha_t)\,\ell^{\text{CE}}_t +
\alpha_t\,\ell^{\text{KD}}_t\big),
\label{eq:ewad}
\end{align}
where $k=10$ sets the sharpness of the gate and $\delta$ is the confidence at which the gate
equals $0.5$. The floor $\eta=0.3$ keeps a CE weight of at least $0.3$ at every token, and the
KL uses $\tau=1$. Unlike plain KD, EWAD mixes the two losses instead of adding them. For
BanglaT5, $\delta=0.5$. For the LLM students, whose top-20 distributions are renormalised,
$\delta$ is the teacher's median top-1 probability on the training set.

\paragraph{CPDP.} CPDP adds a second teacher $T_2$. It asks the student to keep its distances
to the two teachers consistent with the distance between the teachers. With mean-pooled hidden
states projected to a shared space and cosine distance $d$, the added loss is
$\beta\,(d_{ST_1}-d_{ST_2}-d_{T_1T_2})^2$ with $\beta=0.05$. For BanglaT5, the states are
encoder outputs and $T_2$ is an mT5 model \citep{xue-etal-2021-mt5}. For the LLM students, they
are prompt states and $T_2$ is the smaller teacher of the same family.

\paragraph{CHAD.} CHAD gives more KD weight to training examples whose KD signal helps on
validation data. Each example is scored by
\begin{equation}
s(x)=\tfrac{1}{|C|}\textstyle\sum_{c\in C}\cos\big(\nabla\mathcal{L}_{\text{val}}(\theta_c),\,
\nabla\ell^{\text{KD}}(x;\theta_c)\big),
\end{equation}
where $\ell^{\text{KD}}(x)=\sum_t\ell^{\text{KD}}_t$, gradients are with respect to the student's
parameters, and $C$ is a set of student checkpoints, as in TracIn \citep{NEURIPS2020_e6385d39}.
Scores are mapped to weights $h(x)\in[0,1]$ by their rank and renormalised within each batch
$B$, so the gate redistributes the KD weight between examples without reducing it:
\begin{equation}
\mathcal{L}_{\text{CHAD}}=\mathcal{L}_{\text{CE}}+\lambda\,
\frac{\sum_{x\in B}h(x)\,\bar\ell^{\text{KD}}(x)}{\sum_{x\in B}h(x)}.
\end{equation}
For the LLM students, only the LoRA parameters receive gradients, so $s(x)$ is computed exactly
for every example. For BanglaT5, it is computed on a probe of 1{,}500 examples, and a
gradient-boosting regressor on 13 cheap features extends it to the rest. Two cheaper variants,
the \emph{entropy gate} and the \emph{semantic gate}, use only teacher-confidence or
teacher--gold-overlap features. The scores are weakly determined. The regressor's held-out
$R^2$ is at most $0.19$, and for the LLM students the weights from two seeds correlate at only
$0.50$ to $0.81$ (Appendix~\ref{app:gate}).

\paragraph{Summarisation variants.}
\label{sec:summ-methods}
The summarisation pipeline uses two teachers in every KD term and their top-50
log-probabilities. \emph{EWAD-agree} gates each token on how much the two teachers agree, not
on confidence. \emph{EWAD-agree+CPDP-kl} adds a CPDP term on output divergences. The
\emph{agreement-harm gate} weights the KD term by teacher agreement times the teachers'
probability of the gold token, a forward-only stand-in for CHAD. As a control, we also run
EWAD-agree with its gate fixed at its average value ($0.666$). Formulas are in
Appendix~\ref{app:methods}.

\section{Experimental Setup}
\label{sec:setup}

\subsection{Tasks and data}
\label{sec:tasks}

We use six Bangla-centric tasks (statistics in Appendix~\ref{app:data}). Extractive question
answering (\textbf{qa}) uses 5{,}000 question--answer pairs over 316 passages from the two
releases of UDDIPOK \citep{aurpa2023uddipok}, split by passage. Bangla-to-English translation
(\textbf{mt}) uses 5{,}000 sentence pairs sampled from the Bangla--English part of Samanantar
\citep{ramesh-etal-2022-samanantar}. Paraphrase generation (\textbf{pp}) uses 5{,}000 pairs from
BanglaParaphrase \citep{akil-etal-2022-banglaparaphrase}, without identical pairs.
Conversational reply (\textbf{cc}) uses all 574 unique question--answer pairs of a public
Bangla chat dataset \citep{spedrox2024bengalichat}. Sentence-type classification (\textbf{bt})
uses the 3{,}775 unique sentences of BTSD \citep{btsd2023}, with the label generated as one word
(majority class $41\%$). News summarisation (\textbf{summ}) uses 5{,}000 BanSum articles of at
most 1{,}000 tokens \citep{hasan2024bansum}. For the first five tasks, we remove duplicates
after text normalisation and make our own 70/10/20 split with seed 42; summ is split 80/10/10.
With 115 test examples, cc cannot detect effects much below $0.025$ token-F1, so we report it
but do not rely on it.

\subsection{Models}
\label{sec:models}

\paragraph{Students.} All students start from pretrained weights, and a separate student is
trained for each task and method. We use BanglaT5-small (60M parameters, full fine-tuning;
\citealp{bhattacharjee-etal-2023-banglanlg}), gemma-4-E2B \citep{gemma4} and Qwen3-1.7B
\citep{qwen3}. We call the last two the LLM students. They are trained with low-rank adapters
(LoRA; \citealp{hu2022lora}) at rank 16 on the five tasks and with 4-bit quantised LoRA (QLoRA)
at rank 32 on summ.

\paragraph{Teachers.} BanglaT5 is distilled from BanglaT5-base (247M), fully fine-tuned on each
task. Each LLM student has a teacher of its own family: gemma-4-31B-it for gemma and Qwen3-32B
for Qwen. These teachers are fine-tuned with 4-bit LoRA on the same data, and their outputs are
cached once with teacher forcing. The smaller gemma-4-12B and Qwen3-14B serve only as the second
CPDP teacher on the five tasks, not as an ensemble. In summ, both gemma teachers enter every KD
term. Every teacher had to pass a validation check before distillation
(Appendix~\ref{app:hyper}). The teacher beats its CE-trained student in 13 of the 15
student--task pairs, by up to $0.16$ token-F1. The exceptions are gemma on cc and bt, where the
teacher is $0.009$ and $0.016$ behind (Table~\ref{tab:teachers}). A weak teacher therefore
cannot explain the results in 13 of the 15 pairs.

\subsection{Experiments}
\label{sec:experiments}

We use \emph{method} for a training objective (CE, plain KD or a gate) and \emph{comparison} for
the difference between two methods on the same student and task.

\paragraph{Grid.} Each student is distilled on the five tasks with plain KD and three gates. The
LLM students use EWAD-conf, EWAD-conf+CPDP and CHAD; BanglaT5 uses CHAD and the two heuristic
gates. This gives 45 single-seed comparisons of a gated with a plain-KD student. EWAD variants
were also trained for BanglaT5, in a separate implementation that has no plain-KD student
(\S\ref{sec:impl}, Appendix~\ref{app:ewadfw}).

\paragraph{CE baselines.} Every LLM student also has a CE student on every task, trained with
the grid recipe and seed 42.

\paragraph{Seeded study.} The gemma student is trained with seeds 42, 7 and 13 for exactly ten
epochs, with early stopping disabled and an adapter saved after every epoch. From each run we
evaluate two checkpoints: the one that early stopping with patience 3 would have kept, found by
replaying the rule on the run's own validation log, and the final one. Both therefore come from
the same run. On mt we train CE, plain KD, EWAD-conf, EWAD-conf+CPDP-d, CHAD, and CE with label
smoothing (LS) at $\varepsilon\in\{0.05,0.1,0.2\}$. On pp, bt and cc we train CE, plain KD,
EWAD-conf and CHAD. qa, where all students are near the ceiling, was not seeded.

\paragraph{Summarisation.} The gemma student is trained for one epoch in each of the six
summarisation configurations with three seeds. The gold target tokens are identical across all
18 runs.

\subsection{Evaluation}

One script scores every prediction file. On the five tasks, the primary metric is token-F1
after Bangla-aware normalisation, and we also report exact match. On summ, the primary metric
is ROUGE-L \citep{lin-2004-rouge} on whitespace tokens. We also compute sacreBLEU
\citep{post-2018-call}, BERTScore \citep{zhang2020bertscore} with \texttt{xlm-roberta-base},
and the cosine similarity of BGE-M3 embeddings \citep{chen2024bge}, which we call semsim.
Single-seed comparisons use a paired bootstrap \citep{koehn-2004-statistical} and a
permutation test, with 10{,}000 resamples each, and McNemar's test on exact match. Multi-seed
comparisons use Welch's $t$-test on seed means. We apply Holm's correction
\citep{holm1979simple} within each family of comparisons. We test equivalence with two one-sided tests
(TOST; \citealp{lakens2017equivalence}) at $\pm0.032$ token-F1, the largest difference we measured between
two correct CE implementations (\S\ref{sec:impl}), and at a stricter $\pm0.020$.

\section{Gated KD Does Not Beat Plain KD}
\label{sec:parity}

Of the 45 comparisons of a gated with a plain-KD student, one is significant after correction.
It is a loss: CHAD on pp with the gemma student, $-0.013$ token-F1 ($p_{\text{Holm}}=0.024$),
still inside the equivalence margin. No comparison shows a significant gain. Of the 45, 42 are
equivalent to plain KD within $\pm0.032$ and 39 within $\pm0.020$ (Appendix~\ref{app:cells}).
The largest positive effect is CHAD on qa with the Qwen student ($+0.019$, permutation
$p=0.003$). It does not survive correction ($p_{\text{Holm}}=0.093$), and \S\ref{sec:ce} shows
that it reflects a weakness of plain KD. The picture does not depend on the metric. On
BERTScore, one comparison is significant and none is a gain. On semsim, four are significant,
two gains ($+0.008$, $+0.007$) and two losses, and all 45 are equivalent on both metrics.

\section{Seed and Implementation Variation}
\label{sec:floors}

\begin{figure}[t]
\centering
\includegraphics[width=\columnwidth]{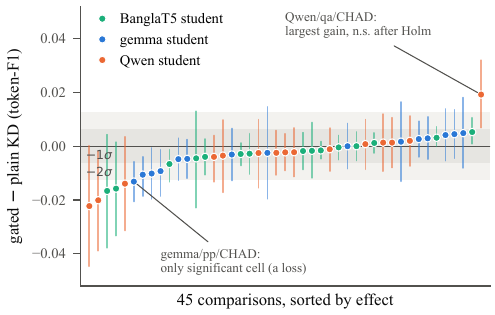}
\caption{The 45 gated-minus-plain-KD token-F1 effects with 95\% bootstrap intervals, against
one and two seed standard deviations measured on mt ($\sigma=0.0063$, shaded bands).}
\label{fig:effects}
\end{figure}

\paragraph{Seed variation.}
\label{sec:seed}
In the seeded study, the pooled within-method standard deviation of token-F1 at the
early-stopped checkpoint is $0.0063$ on mt, $0.0030$ on pp, $0.0068$ on bt and $0.0088$ on cc
(Table~\ref{tab:seeded}). Against the mt value, 34 of the 45 grid effects ($76\%$) lie within
one standard deviation and 38 ($84\%$) within two (Figure~\ref{fig:effects}). For pure noise of
that spread, the expected shares are $68\%$ and $95\%$. For the gemma student, on which the
spread was measured, the counts are 11 and 14 of 15. Three seeds estimate the spread only
roughly, so these counts are descriptive. Still, the grid effects are of the size that
reseeding alone produces.

\paragraph{Implementation variation.}
\label{sec:impl}
The BanglaT5 student has two CE baselines, trained with the recipes of the two codebases in
which the gates were originally built (learning rate $5\times10^{-5}$, batch 16, up to 15
epochs; and $3\times10^{-4}$, batch 32, up to 6 epochs). On identical data and scoring, they
differ significantly on two of five tasks, in opposite directions. On mt the difference is
$+0.032$ ($p_{\text{Holm}}<0.001$); one recipe produces 47 empty outputs in 1{,}000, and
without them the gap is $+0.020$. On qa it is $-0.013$ ($p_{\text{Holm}}=0.008$). We scored the
ten students trained in the EWAD codebase against each CE baseline in turn. For eight of the
ten, switching the baseline changes the sign of the effect, moves it across the $\pm0.032$
margin or changes its significance (Appendix~\ref{app:ewadfw}). The choice between two
reasonable baselines is thus enough to change most of these conclusions, so we report both
wherever two exist.

\paragraph{Software environment.} Re-running the mt seed-42 students on the software stack of
the seeded study moves token-F1 by $0.001$ to $0.004$ ($0.014$ for CHAD, whose scores are
recomputed). The same CE recipe on two different GPUs gives $0.514$ and $0.513$ on pp and
$0.962$ and $0.959$ on bt.

\section{Distillation Does Not Beat Fine-Tuning}
\label{sec:ce}

The grid compares gated with plain KD. A more basic question is whether distillation helps at
all. Table~\ref{tab:ce} compares every distilled LLM student with a CE student trained with the
same recipe. None of the 40 differences is significant after correction, and all lie within
$\pm0.020$ token-F1. Before correction, the confidence interval excludes zero in ten cases,
seven of which favour CE. In the seeded study, no distilled method differs significantly from
CE at the early-stopped checkpoint on any of mt, pp, bt or cc (smallest $p_{\text{Holm}}=0.38$;
Table~\ref{tab:seeded}).

\begin{table}[t]
\centering\small
\resizebox{\columnwidth}{!}{\input{tables/tab_b}}
\caption{Distilled minus CE student, token-F1, single seed (42). None of the 40 differences is
significant after Holm's correction within each student. $^\dagger$Bootstrap 95\% interval
excludes zero before correction.}
\label{tab:ce}
\end{table}

These baselines also explain the largest gated effect in \S\ref{sec:parity}. For Qwen on qa, CE
scores $0.918$, plain KD $0.902$ and CHAD $0.921$. CHAD's advantage of $0.019$ over plain KD is
almost exactly what plain KD lost relative to CE ($0.016$), and CHAD does not differ from CE
($+0.003$, $p_{\text{Holm}}=1.0$). The gate recovers a loss caused by distillation; it does not
improve on CE. Plain KD is likewise below CE in four of the five BanglaT5 student--task pairs
(Table~\ref{tab:teachers}).

\section{Distillation Acts as a Regulariser}
\label{sec:regularise}

\begin{figure}[t]
\centering
\includegraphics[width=\columnwidth]{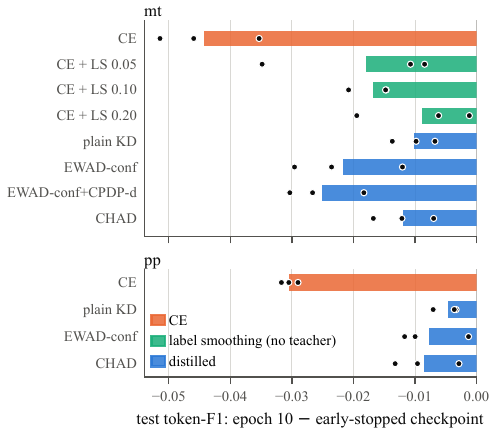}
\caption{Change in test token-F1 from the early-stopped checkpoint to epoch ten, per seed
(points) and on average (bars), on mt and pp. CE degrades most; the distilled methods and
teacher-free label smoothing (LS, mt only) degrade much less.}
\label{fig:degradation}
\end{figure}

Although KD does not improve on CE at the early-stopped checkpoint, it does change how training
proceeds. To measure this, the seeded study compares the early-stopped checkpoint of each run
with its checkpoint after a fixed ten epochs (Table~\ref{tab:seeded},
Figure~\ref{fig:degradation}).

\begin{table*}[t]
\centering\small
\input{tables/tab_seeded}
\caption{Seeded study (gemma student, seeds 42/7/13), test token-F1 as mean $\pm$ standard
deviation over seeds. ``Early stop'' is the checkpoint that patience-3 early stopping keeps,
``epoch 10'' the final checkpoint of the same run, and degradation their difference.
$\Delta$ vs CE and $p_{\text{Holm}}$ are for the early-stopped checkpoint (Welch's test on
seed means). $\sigma$ is the pooled within-method standard deviation. Per-seed values are in
Appendix~\ref{app:seeds}.}
\label{tab:seeded}
\end{table*}

\paragraph{Under a fixed budget, CE overfits most.} Training past the early-stopping point hurts
every method on mt, but it hurts CE most. CE loses $0.044\pm0.008$ token-F1 by epoch ten, while
the distilled methods lose between $0.010$ (plain KD) and $0.025$ (EWAD-conf+CPDP-d). At epoch
ten, every distilled method is therefore $0.020$ to $0.024$ ahead of CE
($p_{\text{Holm}}=0.019$). pp shows the same pattern with less noise. CE loses $0.030\pm0.001$
and the distilled methods $0.005$ to $0.009$, which leaves them $0.019$ to $0.024$ ahead at
epoch ten ($p_{\text{Holm}}=0.004$). The direct test of whether CE degrades more than plain KD
is significant on pp ($p_{\text{Holm}}<0.001$) but not on mt after correction
($p_{\text{Holm}}=0.067$). The gated methods do not degrade differently from plain KD
(Appendix~\ref{app:fixed}). CE's validation score peaks in the first epoch in every seed on
both tasks, whereas on pp plain KD and EWAD-conf keep improving until the fifth to tenth epoch.
On cc and bt, no method degrades: with 402 training examples (cc) or a three-way label (bt), ten
epochs is not a long budget.

\paragraph{Label smoothing reproduces most of the effect.} If the teacher's distribution acts
mainly as a regulariser, smoothing the targets without a teacher should give similar
protection. On mt, we trained CE with label smoothing \citep{muller2019label} at
$\varepsilon\in\{0.05,0.1,0.2\}$ under the same budget. These students degrade by $0.018$,
$0.017$ and $0.009$, against $0.044$ for CE and $0.010$ for plain KD. Label smoothing therefore
removes a share $f=(|d_{\text{CE}}|-|d_{\text{LS}}|)/(|d_{\text{CE}}|-|d_{\text{KD}}|)$ of
$0.77$, $0.80$ and $1.03$ of the CE-to-KD gap, computed on seed means ($0.40$ to $1.58$ per
seed; Appendix~\ref{app:fixed}). At $\varepsilon=0.2$, the degradation is indistinguishable
from that of plain KD ($p_{\text{Holm}}=1.0$). At the early-stopped checkpoint, label smoothing
does not help either: it is $0.006$ to $0.011$ below CE, which is not significant. A single
smoothing constant, with no teacher, thus provides most or all of the protection that
distillation provides here. This also addresses the objection that parity with CE is an
artefact of early stopping. With longer training, KD pulls ahead only because CE degrades, and
every method is then worse than at its own early-stopped checkpoint.

\section{Teacher Quality Barely Reaches the Student}
\label{sec:transfer}

If KD mainly regularises, a much worse teacher should barely change the student. We test this
in one setting, BanglaT5 on qa with a single seed. The pipeline was run twice, once with an
under-trained BanglaT5-base teacher and once with a converged one. The under-trained teacher
comes from an earlier run whose teacher training stopped too soon (ten epochs at learning rate
$5\times10^{-5}$, still improving). The converged teacher was trained at learning rate
$10^{-4}$ for up to 40 epochs. They reach $0.645$ and $0.960$ test exact match. Data, split,
student recipe and CE student are identical, and CHAD's scores are recomputed for each teacher.
Because the CE student uses no teacher, this gives a difference-in-differences comparison. The
weaker teacher is $0.291$ token-F1 worse $[-0.320,-0.262]$, but its students are only $0.002$
(entropy gate) to $0.009$ (semantic gate) worse, an attenuation of about 31 to 1. Only the
semantic-gate difference survives correction on token-F1 ($p_{\text{Holm}}=0.034$), and none
does on BERTScore or semsim (Appendix~\ref{app:weak}). This is the pattern the regularisation
account predicts. Because it rests on one task, one student and one seed, we treat it as
supporting evidence, not as a test on its own.

\section{Summarisation}
\label{sec:summ}

The gates were designed for summarisation, where outputs are long and teacher distributions are
less peaked, so this is where they should help most.

\begin{table}[t]
\centering
\resizebox{\columnwidth}{!}{\input{tables/tab_summ}}
\caption{Summarisation (gemma student, seeds 42/7/13), ROUGE-L as mean $\pm$ sd; BERTScore and
semsim are seed means, and ``words'' the mean output length (references: 73.1). $\Delta$ and
$p_{\text{Holm}}$ are against CE (Welch's test on seed means, Holm over eight contrasts).
All contrasts on all metrics are in Appendix~\ref{app:summ}.}
\label{tab:summ}
\end{table}

\paragraph{No configuration beats CE.} CE has the highest ROUGE-L ($0.279\pm0.002$), BERTScore
and semsim (Table~\ref{tab:summ}). Every distilled configuration is below CE on ROUGE-L. The
difference is significant after correction for plain KD ($-0.019$), the agreement-harm gate
($-0.011$) and the constant gate ($-0.033$), and not significant for EWAD-agree ($-0.006$) or
EWAD-agree+CPDP-kl ($-0.017$). EWAD-agree does beat plain KD ($+0.013$,
$p_{\text{Holm}}=0.027$). Measured against plain KD, the gate therefore shows a gain; measured
against CE, it does not.

\paragraph{The learned gate beats a constant gate.} Fixing EWAD-agree's gate at its average
value keeps the average KD weight but removes the variation between tokens. The learned gate
beats this constant by $0.027$ ROUGE-L ($p_{\text{Holm}}=0.005$). At that average, both put
about eight times more weight on the KD term than on CE, compared with two times for plain KD.
Spread uniformly, this much KD gives the worst configuration. The learned gate concentrates it
on tokens where the teachers agree and returns the rest of the weight to CE. This brings the
student back towards CE, but not past it.

\paragraph{Length and metrics.} CE's summaries ($69.2$ words) are close to the references in
length ($73.1$); EWAD-agree writes $75.8$ words and plain KD $56.4$. Truncating every output to
47 words shrinks all ROUGE-L gaps without changing any sign (EWAD-agree minus CE moves from
$-0.006$ to $-0.004$). ROUGE-L, BERTScore and semsim all rank CE first and the constant gate
last (Spearman $\rho=1.0$ between ROUGE-L and semsim, $0.71$ between BERTScore and either). On
the five tasks, all three metrics agree on the sign of 10 of the 12 seeded CE-versus-KD
differences, and the other two are smaller than the seed standard deviation
(Appendix~\ref{app:metrics}).

\section{Relation to the Original Results}
\label{sec:replication}

\citet{sumit2026when} report that CHAD improves ROUGE-L by $0.018$ and EWAD+CPDP by $0.022$
over a CE student of the same architecture on BanSum. Our study does not repeat that setting.
They distil BanglaT5-small (60M) from BanglaT5 (247M) on about 141{,}000 BanSum articles,
roughly 28 times our summarisation training set. Our summarisation student is a 2B model,
distilled from 31B and 12B teachers through their top 50 tokens, and our BanglaT5 students see
at most 5{,}000 examples of other tasks. Our results therefore do not show that their gains are
wrong in their setting. They show that the gains do not carry over to ours, and they identify
what a test in their setting would need.

Three of our controls are relevant. \emph{Seeds:} gains of $0.018$ and $0.022$ are at least
three times the summarisation seed standard deviation we measure ($0.002$ to $0.006$), so seed
noise alone is an unlikely explanation. However, their main configurations were run once, and
their multi-seed check covered a 20{,}000-example subset, not those configurations.
\emph{Baseline:} two correct CE recipes for our BanglaT5 student differ by up to $0.032$
token-F1 and change the conclusion for eight of ten students (\S\ref{sec:impl}). A gain over a
single CE run therefore depends on that run's recipe. \emph{Constant gate:} in our summarisation
experiments, the learned gate's advantage comes from where it places the KD weight, and it
recovers towards CE without passing it (\S\ref{sec:summ}). A gain over CE needs this control
before it can be attributed to the gate. Checkpoint selection is comparable in the two studies
(300 validation articles there, 150 or 200 here) and does not explain the difference.

Our configuration closest to theirs is BanglaT5 with CPDP. It gives our largest single-seed
gain: $+0.048$ token-F1 on mt over the CE student of its own codebase ($p_{\text{Holm}}<0.001$).
That CE student is the one with 47 empty outputs. Against the other CE student, the gain is
$+0.016$ $[+0.007,+0.026]$, inside the stricter equivalence margin. Gating may matter more with
far more data, a smaller capacity gap or full teacher distributions. A multi-seed test against
both baselines and a constant gate in that setting is the natural next step.

\section{Discussion and Conclusion}
\label{sec:discussion}

Taken together, the results are consistent with one explanation. In this setting, the quality
of the teacher barely reaches the student (\S\ref{sec:transfer}, one task and one seed). What
distillation provides is protection against overfitting, and this protection does not need a
teacher: one label-smoothing constant reproduces most of it (\S\ref{sec:regularise}). A gate
applied to such a signal can change how much regularisation is applied, but it has little
teacher knowledge to select. This explanation predicts parity with plain KD
(\S\ref{sec:parity}) at the size of seed noise (\S\ref{sec:floors}), no advantage over CE at the
early-stopped checkpoint (\S\ref{sec:ce}), and a learned gate whose measurable benefit is to
repair damage that distillation itself causes (\S\ref{sec:summ}).

For low-resource generation at this scale, CE fine-tuning with early stopping is a strong
baseline that no form of distillation beat in our experiments. A claim that a gate improves
distillation should be supported by four controls: a comparison with a CE student, not only
with plain KD; a constant gate of the same average weight; at least three seeds; and, where two
reasonable baseline implementations exist, both of them. Each of these controls changed a
conclusion in this study. We do not claim that distillation is useless in general. For
Bangla-centric tasks, small students and teachers fine-tuned on the same data, its effect is
mainly a regularisation that is available more cheaply.

\section*{Limitations}

\paragraph{Seeds.} The 45-comparison grid and the CE baselines of the LLM students use a single
seed. The seeded study uses three seeds for the gemma student on four of the five tasks (not
qa), and the summarisation study uses three seeds. Three seeds estimate the seed variation
only roughly, and Holm's correction is strict at this sample size; we therefore report
uncorrected and corrected $p$-values and give every per-seed value in
Appendix~\ref{app:seeds}. The cc test set ($n=115$) is too small to detect effects below about
$0.025$ token-F1.

\paragraph{Scope of the regularisation evidence.} The fixed-budget effect appears on mt and pp;
on cc and bt no method overfits within ten epochs, so those tasks neither support nor contradict
it. The label-smoothing control was run on mt only.

\paragraph{Weak-teacher comparison.} The teacher-quality comparison (\S\ref{sec:transfer}) comes
from one task, one student and one seed, and the weak teacher was under-trained rather than
deliberately corrupted. Because CHAD's scores and gates were recomputed for each teacher, the
gated students differ in more than the teacher.

\paragraph{Evaluation.} All results use automatic metrics; we did not run a human or LLM-judge
evaluation. A human study limited to summarisation could not support claims across the six
tasks, and on every task the lexical, contextual and embedding-based metrics we report agree
on the direction of the effects wherever there is an effect larger than the seed variation
(\S\ref{sec:summ}, Appendix~\ref{app:metrics}). Whether human readers would prefer CE or
distilled summaries remains open.

\paragraph{Generalisation.} All tasks are Bangla-centric, students have 60M to about 2B
parameters, and teachers are fine-tuned on the same data as the students. The LLM students see
only the top 20 (five tasks) or top 50 (summarisation) teacher tokens. We do not reproduce the
original BanSum setting of \citet{sumit2026when}, which uses far more data and a smaller
capacity gap (\S\ref{sec:replication}). Settings with much larger capacity gaps, out-of-domain
teachers, much more training data or full teacher distributions may behave differently; the
weak-teacher and label-smoothing designs used here can be reused to test them. We test two gate
families and do not claim that other ways of selecting the teacher signal would fail.

\section*{Ethics Statement}

All datasets are publicly available and were used in accordance with their licences; we
collected no new data and recruited no human participants. Bangla news and conversation data
may contain social biases that generated text can reproduce, and deployed systems should
include human review. Experiments ran on single consumer GPUs. Code, data splits, per-example
predictions and scores, and pinned environment specifications will be released publicly; the data splits keep the licences of the original
datasets. AI assistants were used for
coding and writing support.

\bibliography{refs}

\appendix

\section{Method Details}
\label{app:methods}

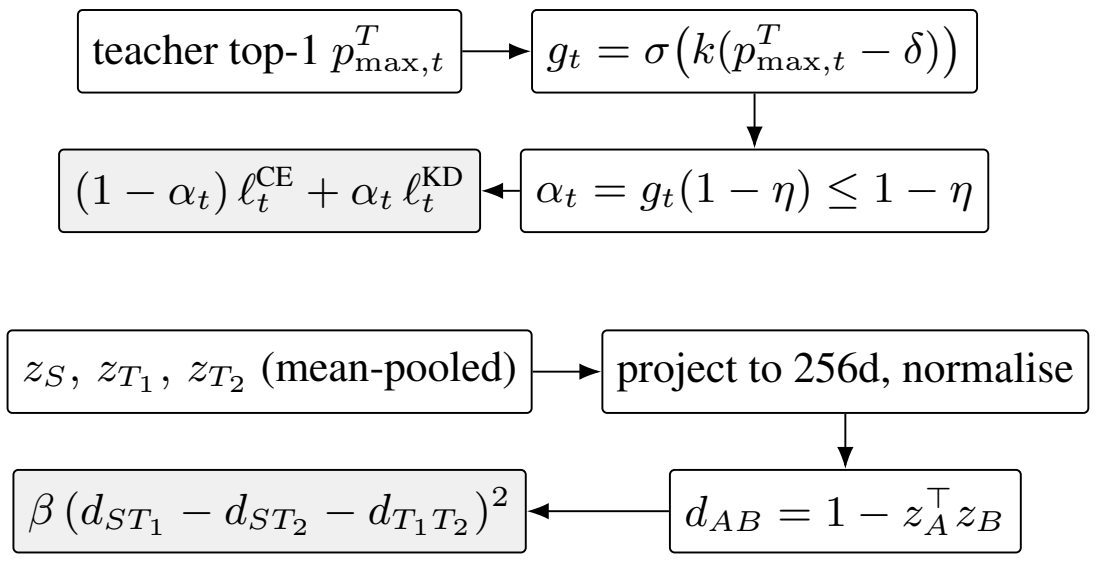
\begin{figure}[t]
\centering
\resizebox{0.93\columnwidth}{!}{%
\begin{tikzpicture}[
  font=\small,
  box/.style={draw, rounded corners=1pt, align=center, inner sep=3pt, minimum height=6mm},
  loss/.style={box, fill=black!6},
  ar/.style={-{Latex[length=1.6mm]}, thin}]
\node[box] (pmax) {teacher top-1 $p^T_{\max,t}$};
\node[box, right=5mm of pmax] (gate) {$g_t=\sigma\!\left(k(p^T_{\max,t}-\delta)\right)$};
\node[box, below=4mm of gate] (alpha) {$\alpha_t=g_t(1-\eta)\le 1-\eta$};
\node[loss, below=4mm of pmax] (mix) {$(1-\alpha_t)\,\ell^{\text{CE}}_t+\alpha_t\,\ell^{\text{KD}}_t$};
\draw[ar] (pmax) -- (gate);
\draw[ar] (gate) -- (alpha);
\draw[ar] (alpha) -- (mix);
\node[box, below=7mm of mix] (z) {$z_S,\,z_{T_1},\,z_{T_2}$ (mean-pooled)};
\node[box, right=5mm of z] (proj) {project to 256d, normalise};
\node[box, below=4mm of proj] (dist) {$d_{AB}=1-z_A^{\top}z_B$};
\node[loss, below=4mm of z] (cpdp) {$\beta\,(d_{ST_1}-d_{ST_2}-d_{T_1T_2})^2$};
\draw[ar] (z) -- (proj);
\draw[ar] (proj) -- (dist);
\draw[ar] (dist) -- (cpdp);
\end{tikzpicture}}
\caption{EWAD-conf (top) and CPDP (bottom). The gate turns the teacher's confidence at a token
into the weight $\alpha_t$ that the KD loss receives there, and the floor $\eta$ reserves weight
for the gold label. CPDP adds a second teacher through distances between projected hidden
states, which stays defined when $T_2$ has a different vocabulary. The two terms are summed into
one objective.}
\label{fig:ewad}
\end{figure}

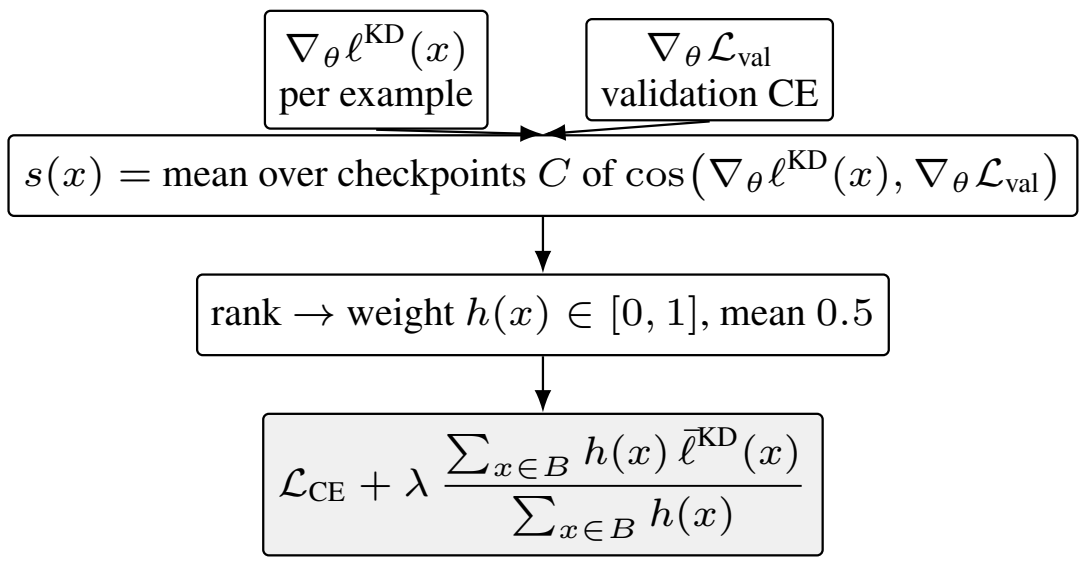
\begin{figure}[t]
\centering
\resizebox{0.92\columnwidth}{!}{%
\begin{tikzpicture}[
  font=\scriptsize,
  box/.style={draw, rounded corners=1pt, align=center, inner sep=2.5pt, minimum height=5mm},
  loss/.style={box, fill=black!6},
  ar/.style={-{Latex[length=1.4mm]}, thin}]
\node[box] (gkd) {$\nabla_\theta\ell^{\text{KD}}(x)$\\ per example};
\node[box, right=6mm of gkd] (gval) {$\nabla_\theta\mathcal{L}_{\text{val}}$\\ validation CE};
\node[box, below=4mm of $(gkd)!0.5!(gval)$] (cos) {$s(x)=$ mean over checkpoints $C$ of
  $\cos\!\big(\nabla_\theta\ell^{\text{KD}}(x),\nabla_\theta\mathcal{L}_{\text{val}}\big)$};
\node[box, below=3.5mm of cos] (rank) {rank $\rightarrow$ weight $h(x)\in[0,1]$, mean $0.5$};
\node[loss, below=3.5mm of rank] (loss)
  {$\mathcal{L}_{\text{CE}}+\lambda\,\dfrac{\sum_{x\in B}h(x)\,\bar\ell^{\text{KD}}(x)}{\sum_{x\in B}h(x)}$};
\draw[ar] (gkd.south) -- (cos.north);
\draw[ar] (gval.south) -- (cos.north);
\draw[ar] (cos) -- (rank);
\draw[ar] (rank) -- (loss);
\end{tikzpicture}}
\caption{CHAD. Each training example is scored by how well its KD gradient aligns with the
validation gradient, averaged over student checkpoints, and the scores become per-example
weights by their rank. The weights are renormalised within the batch, so CHAD redistributes the
KD term between examples rather than reducing it. For the BanglaT5 student, the scores are computed
for a probe of 1{,}500 examples and a regressor generalises them to the rest.}
\label{fig:chad}
\end{figure}

Figures~\ref{fig:ewad} and~\ref{fig:chad} show EWAD-conf, CPDP and CHAD as used on the five
tasks.

\paragraph{CPDP variants.} For the BanglaT5 student (CPDP-geo), the hidden states are encoder outputs,
mean-pooled with the attention mask, each passed through its own linear projection to 256
dimensions (orthogonally initialised and trained with the student) and L2-normalised; $T_2$ is
an mT5 model fine-tuned on the same task, whose vocabulary differs from the student's. For the LLM students (CPDP-d), the same loss is applied to mean-pooled hidden states of the prompt, taken
from the cached teacher passes, with $T_2$ the smaller teacher of the same family. These
teachers share the student's vocabulary, so a KL to $T_2$ would also be possible; we keep the
representation-distance form so that the method matches its original design.

\paragraph{CHAD for the BanglaT5 student.} The probe contains 1{,}500 training examples, the
validation gradient is taken over 300 validation examples, and $C$ is the pretrained student
and up to three checkpoints of the CE-trained student. A gradient-boosting regressor predicts
$s(x)$ from 13 features (teacher entropy, top-1 probability and top-2 margin; student CE loss
and entropy; teacher--student KL; teacher--gold ROUGE-L; and length, compression and novelty
features) and scores every training example. The entropy gate uses only teacher entropy, top-1
probability, top-2 margin and source and target length; the semantic gate uses only
teacher--gold ROUGE-L and the two lengths. For the LLM students, $C$ is the plain-KD student's
adapters after its first three epochs, and the scores are recomputed for every seed.

\paragraph{Summarisation variants.} Plain KD uses the fixed mixture
$0.7\,\mathrm{KL}_1+0.3\,\mathrm{KL}_2$ of the two teachers. \emph{EWAD-agree} measures teacher
agreement as $A_t=\sum_v\min(p^{T_1}_v,p^{T_2}_v)$, a bounded proxy for one minus the
Jensen--Shannon divergence, gates with $g_t=\sigma(k(A_t-0.5))$, and uses the loss
$g_t\,\tau^2(w_1\mathrm{KL}_1+w_2\mathrm{KL}_2)+(1-g_t)\,\ell^{\text{CE}}_t$, where the weights
$w_i$ are a softmax over the teachers' normalised confidence. \emph{EWAD-agree+CPDP-kl} adds
$\mu\,\big((\mathrm{KL}_1-\mathrm{KL}_2)/H(p^S)-\mathrm{KL}(p^{T_1}\|p^{T_2})\big)^2$ with
$\mu=0.05$. The \emph{agreement-harm gate} uses
$\ell^{\text{CE}}_t+h_t\,\tau^2(w_1\mathrm{KL}_1+w_2\mathrm{KL}_2)$ with
$h_t=\eta+(1-\eta)\,A_t\,m_t$, where $m_t$ is the teachers' average probability of the gold token
and $\eta=0.3$. The constant-gate control sets $g_t=0.666$, the gate's average over the training
targets. With $\tau=2$, plain KD weights KD to CE at $2:1$ and the constant gate at about $8:1$.

\section{Data and Teachers}
\label{app:data}

\begin{table}[t]
\centering
\resizebox{\columnwidth}{!}{\input{tables/tab_tasks}}
\caption{Task statistics. ``ref.\ words'' is the mean length of the test references in
whitespace-separated words.}
\label{tab:tasks}
\end{table}

\begin{table}[t]
\centering
\resizebox{\columnwidth}{!}{\input{tables/tab_teachers}}
\caption{Test token-F1 (accuracy for bt) of the teacher (T), the CE-trained student (CE)
and the plain-KD student (KD). $^{\ast}$Teacher below the CE student. Single seed; the gemma
mt CE value is seed 42 of the seeded study.}
\label{tab:teachers}
\end{table}

Table~\ref{tab:tasks} gives the size of every split and Table~\ref{tab:teachers} the quality of
every teacher relative to the students it distils into.

\section{All Gated-versus-Plain-KD Comparisons}
\label{app:cells}

\begin{table}[t]
\centering\small
\resizebox{\columnwidth}{!}{\input{tables/tab_parity}}
\caption{Gated versus plain KD on token-F1, single seed. ``sig.'' counts comparisons
significant after Holm's correction within each group, ``gains'' those that are significant
and favour the gate, and ``equiv.'' those shown equivalent by TOST at each margin.}
\label{tab:parity}
\end{table}

Table~\ref{tab:parity} summarises the comparisons of \S\ref{sec:parity}, and
Table~\ref{tab:cells45} lists each one with its paired-bootstrap 95\% interval, permutation
$p$-value, Holm-corrected $p$-value (within each group) and TOST equivalence at $\pm0.032$.

\begin{table*}[t]
\centering\small
\input{tables/tab_cells45}
\caption{All 45 gated-minus-plain-KD comparisons, token-F1, single seed. ``equiv.'' marks
equivalence at $\pm0.032$ by TOST.}
\label{tab:cells45}
\end{table*}

\section{EWAD for the BanglaT5 Student}
\label{app:ewadfw}

EWAD-conf and EWAD-conf+CPDP-geo were trained for the BanglaT5 student in a separate
implementation (learning rate $3\times10^{-4}$, effective batch 32, up to 6 epochs, early
stopping on validation ROUGE-L) without a plain-KD student, so they are not part of the grid.
Table~\ref{tab:ewadfw} compares each with the CE student of the same implementation and with
the CE student of the other implementation (\S\ref{sec:impl}).

\begin{table*}[t]
\centering\small
\input{tables/tab_ewadfw}
\caption{EWAD variants for the BanglaT5 student, single seed, token-F1. $p_{\text{Holm}}$ is for the
comparison with the CE student of the same implementation (Holm over the ten comparisons).}
\label{tab:ewadfw}
\end{table*}

\section{Per-Seed Values}
\label{app:seeds}

Tables~\ref{tab:perseedA} and~\ref{tab:perseedC} give the value of every individual run in the
seeded study and the summarisation study.

\begin{table*}[t]
\centering\small
\input{tables/tab_perseed_A}
\caption{Seeded study, test token-F1 of every run at the early-stopped and final checkpoints,
and the epoch (0-based) that early stopping selected in each seed.}
\label{tab:perseedA}
\end{table*}

\begin{table}[t]
\centering
\resizebox{\columnwidth}{!}{\input{tables/tab_perseed_C}}
\caption{Summarisation, every run.}
\label{tab:perseedC}
\end{table}

\section{Fixed-Budget Contrasts}
\label{app:fixed}

\begin{table}[t]
\centering
\resizebox{\columnwidth}{!}{\input{tables/tab_fixed_vs}}
\caption{Fixed ten-epoch budget. Left: epoch-ten token-F1 minus CE. Right: degradation minus
that of plain KD (negative means the method degrades more). Welch's test on seed means, Holm
within each task.}
\label{tab:fixedvs}
\end{table}

\begin{table}[t]
\centering
\resizebox{\columnwidth}{!}{\input{tables/tab_ls}}
\caption{Label smoothing on mt (three seeds). Degradation is epoch-ten minus early-stopped
token-F1 (mean $\pm$ sd). $f$ is the share of the CE-to-plain-KD degradation gap removed. The
last two columns compare the early-stopped checkpoint with CE (Welch's test, uncorrected).}
\label{tab:ls}
\end{table}

\begin{figure}[t]
\centering
\includegraphics[width=\columnwidth]{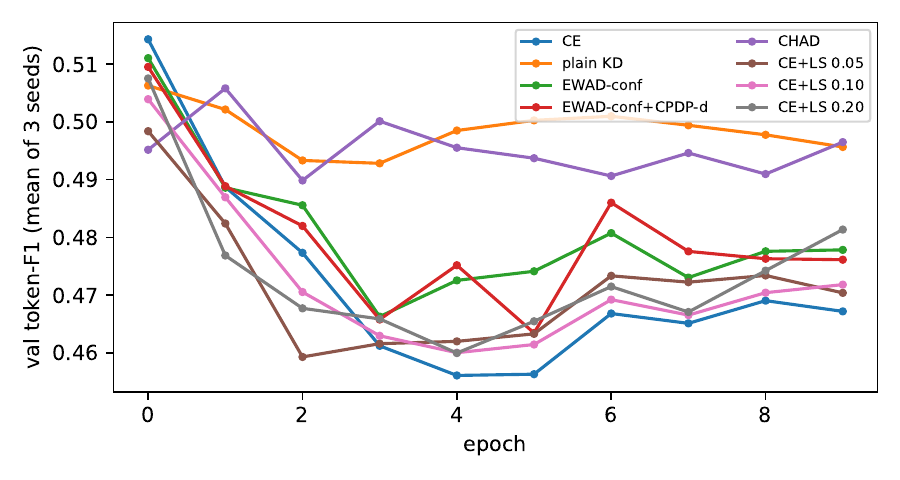}
\caption{Validation token-F1 on mt over ten epochs, mean of three seeds.}
\label{fig:traj}
\end{figure}

Tables~\ref{tab:fixedvs} and~\ref{tab:ls} give the contrasts behind \S\ref{sec:regularise}.
Figure~\ref{fig:traj} shows the validation token-F1 of every mt method over the ten epochs,
averaged over seeds.

\section{CHAD Score Quality}
\label{app:gate}

\begin{table}[t]
\centering\small
\resizebox{\columnwidth}{!}{\input{tables/tab_gatequality}}
\caption{CHAD score quality. Upper rows: the BanglaT5 student, where a regressor predicts the
score; ``helpful share'' is the fraction of probe examples with $s(x)>0$, and $R^2$ and AUC
are on held-out probe examples. Last row: gemma student, where scores are exact; range of the
Spearman correlation of the CHAD weights between pairs of seeds.}
\label{tab:gatequality}
\end{table}

Table~\ref{tab:gatequality} gives the diagnostics summarised in \S\ref{sec:methods}: how well
CHAD's score can be predicted for the BanglaT5 student, and how stable the exact scores are across
seeds for the gemma student. A gate whose held-out $R^2$ is at most $0.19$, and whose weights
correlate at $0.50$ to $0.81$ between seeds, is close to a noisy reweighting of plain KD. The
share of examples with $s(x)>0$ ranges from $21\%$ (pp) to $93\%$ (qa), against about half on
BanSum in \citet{sumit2026when}.

\section{Weak versus Converged Teacher}
\label{app:weak}

Table~\ref{tab:weak} gives the full comparison behind \S\ref{sec:transfer}: for the teacher
and for each student, the difference between the run with the under-trained teacher and the run
with the converged one, on the same test set and with the CE student shared between the two
conditions. The teacher moves by $-0.29$ token-F1 and no student by more than $0.009$; only the
semantic gate's drop survives Holm's correction on token-F1, and the BERTScore column shows the
same ordering at a tenth of the scale.

\begin{table}[t]
\centering
\resizebox{\columnwidth}{!}{\input{tables/tab_weak}}
\caption{Under-trained minus converged teacher on qa (BanglaT5 student, single seed), for the teacher
and each student. $p_{\text{Holm}}$ is over the four distilled students.}
\label{tab:weak}
\end{table}

\section{Summarisation Contrasts}
\label{app:summ}

Table~\ref{tab:summcontrasts} lists all eight contrasts of \S\ref{sec:summ} on each of the
three metrics, rather than only the comparisons with CE reported in the main text. The three
metrics agree on the two effects that matter for the argument: the learned gate beats the
constant gate of the same average weight, and no configuration beats CE. They disagree on
magnitude only, with BERTScore compressing every difference into a range of $0.006$.

\begin{table*}[t]
\centering\small
\input{tables/tab_summ_contrasts}
\caption{Summarisation contrasts on all three metrics (Welch's test on seed means, Holm over the
eight contrasts of each metric).}
\label{tab:summcontrasts}
\end{table*}

\section{Agreement Between Metrics}
\label{app:metrics}

Because this study reports no human or LLM-judge evaluation, Table~\ref{tab:metricagree} checks
whether the automatic metrics could be hiding an effect from each other. For every distilled method
of the seeded study it gives the difference from CE on token-F1, BERTScore and semantic
similarity. In 10 of the 12 comparisons all three metrics agree on the sign; in the two
exceptions the token-F1 difference is smaller than that task's seed standard deviation, so the
disagreement is within noise rather than between measures.

\begin{table}[t]
\centering
\resizebox{\columnwidth}{!}{\input{tables/tab_metric_agreement}}
\caption{Seeded study: distilled method minus CE at the early-stopped checkpoint, on three metrics
(seed means). The last column marks whether the token-F1 difference is below the task's seed
standard deviation.}
\label{tab:metricagree}
\end{table}

\section{Hyperparameters and Environment}
\label{app:hyper}

\begin{table}[t]
\centering\small
\setlength{\tabcolsep}{3pt}
\begin{tabular}{@{}lll@{}}
\toprule
setting & LLM students & BanglaT5 student \\
\midrule
adaptation & LoRA $r{=}16$, bf16 & full fine-tuning \\
optimiser & AdamW & AdamW \\
learning rate & $2\times10^{-4}$ & $5\times10^{-5}$ \\
effective batch & $4\times4=16$ & $8\times2=16$ \\
epochs / patience & 10 / 3 & 15 / 6 evaluations \\
KD $\lambda$ / $\tau$ & 0.5 / 2.0 & 0.5 / 2.0 \\
EWAD $k$ / $\delta$ / $\eta$ & 10 / median / 0.3 & 10 / 0.5 / 0.3 \\
CPDP weight & 0.05 & 0.05 \\
teacher distribution & cached top-20 & full vocabulary \\
decoding & greedy & beam 4 \\
\bottomrule
\end{tabular}
\caption{Core hyperparameters on the five tasks. qa uses batch $2\times8$ for the LLM students.
Source/target token budgets: qa 512/32, mt, pp and cc 128/128, bt 128/16.}
\label{tab:hyper}
\end{table}

\paragraph{Teachers.} The LLM teachers are fine-tuned with 4-bit LoRA at learning rate $10^{-4}$
and must reach validation token-F1 of at least $0.40$ before their outputs are cached. The
BanglaT5 teacher is fully fine-tuned at learning rate $10^{-4}$ for up to 40 epochs and must
reach, on validation, exact match $\geq0.80$ (qa) or token-F1 $\geq0.25$ (mt, pp), $\geq0.45$
(bt, above the $0.41$ majority class) or $\geq0.15$ (cc).

\paragraph{CHAD.} BanglaT5 student: probe of 1{,}500 training and 300 validation examples,
gradient-boosting gate with 300 trees of depth 4 and learning rate 0.05, trained on 1{,}200
probe examples. LLM students: exact scores for every training example over the LoRA parameters.

\paragraph{Summarisation.} gemma-4-E2B student with 4-bit QLoRA (rank 32, $\alpha=64$), batch
$2\times8$, learning rate $2\times10^{-4}$ with cosine schedule, one epoch, $\lambda=0.5$,
$\tau=2$; source up to 1{,}024 and target up to 256 tokens; teachers gemma-4-31B-it and
gemma-4-12B through cached top-50 teacher-forced log-probabilities; greedy decoding with up to
200 new tokens.

\paragraph{Environment.} torch 2.10.0 with CUDA 12.8, transformers 5.14.1, bitsandbytes 0.50.0,
bert\_score 0.3.12, sacreBLEU 2.6.0. The mt seeded runs, the CE baselines and summarisation ran
on an RTX 5090 (32\,GB); the pp, bt and cc seeded runs on an RTX 5070\,Ti (16\,GB) with gradient
checkpointing, which changes memory use but not the computed gradients. Model revisions are
pinned on every load.

\paragraph{Reproducing the results.} The released material contains the data splits, every prediction
file, the per-example scores, the run configurations and the scripts that produce each table and figure
from them: \texttt{analysis/score\_all.py} (scoring), \texttt{analyze\_grid.py}
(\S\ref{sec:parity}, \S\ref{sec:floors}, \S\ref{sec:transfer}),
\texttt{analyze\_A\_mechanism.py} (\S\ref{sec:regularise}), \texttt{analyze\_B\_ce.py}
(\S\ref{sec:ce}), \texttt{analyze\_C\_summ.py} (\S\ref{sec:summ}),
\texttt{metric\_agreement.py} (Appendix~\ref{app:metrics}), and \texttt{paper/make\_tables.py},
which writes every table in this paper. \texttt{acceptance\_check.py} verifies that a new
environment reproduces the archived numbers.

\end{document}

%% file: tables/tab_b.tex
\begin{tabular}{@{}llrrrrr@{}}
\toprule
student & task & CE F1 & plain KD & EWAD-conf & +CPDP & CHAD \\
\midrule
\multirow{5}{*}{gemma} & qa & 0.9532 & $+0.0002$ & $+0.0046$ & $+0.0002$ & $+0.0031$ \\
 & mt & 0.5161 & $-0.0023$ & $-0.0054$ & $+0.0019$ & $-0.0124^{\dagger}$ \\
 & pp & 0.5144 & $+0.0057$ & $+0.0009$ & $+0.0010$ & $-0.0075^{\dagger}$ \\
 & cc & 0.7168 & $+0.0116$ & $+0.0164$ & $+0.0132$ & $+0.0091$ \\
 & bt & 0.9616 & $-0.0053$ & $-0.0146^{\dagger}$ & $-0.0159^{\dagger}$ & $-0.0026$ \\
\midrule
\multirow{5}{*}{Qwen} & qa & 0.9181 & $-0.0162^{\dagger}$ & $-0.0197^{\dagger}$ & $-0.0187^{\dagger}$ & $+0.0030$ \\
 & mt & 0.4370 & $+0.0012$ & $+0.0032$ & $+0.0020$ & $-0.0010$ \\
 & pp & 0.4431 & $+0.0041$ & $+0.0017$ & $+0.0030$ & $+0.0018$ \\
 & cc & 0.6718 & $+0.0127$ & $-0.0096$ & $-0.0013$ & $-0.0074$ \\
 & bt & 0.9470 & $+0.0119^{\dagger}$ & $+0.0079$ & $+0.0132^{\dagger}$ & $+0.0132^{\dagger}$ \\
\bottomrule
\end{tabular}

%% file: tables/tab_seeded.tex
\begin{tabular}{@{}llrrrrr@{}}
\toprule
task & method & F1 at early stop & $\Delta$ vs CE & $p_{\text{Holm}}$ & F1 at epoch 10 & degradation \\
\midrule
\multirow{5}{*}{\shortstack[l]{mt\\($n{=}1000$,\\ $\sigma{=}0.0063$)}} & CE & 0.5216 $\pm$ 0.0048 & -- & -- & 0.4774 & $-0.0442$ $\pm$ 0.0082 \\
 & plain KD & 0.5114 $\pm$ 0.0009 & $-0.0102$ & 0.430 & 0.5013 & $-0.0101$ $\pm$ 0.0035 \\
 & EWAD-conf & 0.5213 $\pm$ 0.0109 & $-0.0003$ & 1.000 & 0.4996 & $-0.0217$ $\pm$ 0.0089 \\
 & EWAD-conf+CPDP-d & 0.5220 $\pm$ 0.0067 & $+0.0004$ & 1.000 & 0.4969 & $-0.0251$ $\pm$ 0.0062 \\
 & CHAD & 0.5135 $\pm$ 0.0038 & $-0.0081$ & 0.505 & 0.5016 & $-0.0120$ $\pm$ 0.0049 \\
\midrule
\multirow{4}{*}{\shortstack[l]{pp\\($n{=}1000$,\\ $\sigma{=}0.0030$)}} & CE & 0.5147 $\pm$ 0.0018 & -- & -- & 0.4843 & $-0.0304$ $\pm$ 0.0013 \\
 & plain KD & 0.5101 $\pm$ 0.0033 & $-0.0045$ & 0.387 & 0.5055 & $-0.0047$ $\pm$ 0.0021 \\
 & EWAD-conf & 0.5162 $\pm$ 0.0042 & $+0.0015$ & 0.612 & 0.5085 & $-0.0077$ $\pm$ 0.0056 \\
 & CHAD & 0.5118 $\pm$ 0.0020 & $-0.0029$ & 0.387 & 0.5032 & $-0.0086$ $\pm$ 0.0053 \\
\midrule
\multirow{4}{*}{\shortstack[l]{bt\\($n{=}755$,\\ $\sigma{=}0.0068$)}} & CE & 0.9558 $\pm$ 0.0043 & -- & -- & 0.9558 & $+0.0000$ $\pm$ 0.0066 \\
 & plain KD & 0.9479 $\pm$ 0.0073 & $-0.0079$ & 0.391 & 0.9563 & $+0.0084$ $\pm$ 0.0061 \\
 & EWAD-conf & 0.9501 $\pm$ 0.0099 & $-0.0057$ & 0.432 & 0.9554 & $+0.0053$ $\pm$ 0.0093 \\
 & CHAD & 0.9497 $\pm$ 0.0035 & $-0.0062$ & 0.380 & 0.9572 & $+0.0075$ $\pm$ 0.0028 \\
\midrule
\multirow{4}{*}{\shortstack[l]{cc\\($n{=}115$,\\ $\sigma{=}0.0088$)}} & CE & 0.7177 $\pm$ 0.0126 & -- & -- & 0.7184 & $+0.0007$ $\pm$ 0.0114 \\
 & plain KD & 0.7115 $\pm$ 0.0106 & $-0.0062$ & 1.000 & 0.7223 & $+0.0108$ $\pm$ 0.0092 \\
 & EWAD-conf & 0.7241 $\pm$ 0.0054 & $+0.0064$ & 1.000 & 0.7253 & $+0.0012$ $\pm$ 0.0106 \\
 & CHAD & 0.7194 $\pm$ 0.0029 & $+0.0017$ & 1.000 & 0.7158 & $-0.0037$ $\pm$ 0.0018 \\
\bottomrule
\end{tabular}

%% file: tables/tab_summ.tex
\begin{tabular}{@{}lrrrrrr@{}}
\toprule
method & ROUGE-L & $\Delta$ vs CE & $p_{\text{Holm}}$ & BERTScore & semsim & words \\
\midrule
CE & 0.2787 $\pm$ 0.0023 & -- & -- & 0.8933 & 0.8487 & 69.2 \\
EWAD-agree & 0.2731 $\pm$ 0.0020 & $-0.0056$ & 0.095 & 0.8922 & 0.8484 & 75.8 \\
agreement-harm gate & 0.2679 $\pm$ 0.0031 & $-0.0108$ & 0.040 & 0.8919 & 0.8442 & 66.4 \\
EWAD-agree+CPDP-kl & 0.2621 $\pm$ 0.0062 & $-0.0166$ & 0.095 & 0.8929 & 0.8432 & 60.2 \\
plain KD & 0.2602 $\pm$ 0.0030 & $-0.0185$ & 0.008 & 0.8920 & 0.8406 & 56.4 \\
EWAD-agree, constant gate & 0.2458 $\pm$ 0.0033 & $-0.0329$ & 0.002 & 0.8870 & 0.8350 & 61.3 \\
\bottomrule
\end{tabular}

%% file: tables/tab_tasks.tex
\begin{tabular}{@{}lrrrrl@{}}
\toprule
task & train & val & test & ref.\ words & type \\
\midrule
qa & 3606 & 488 & 906 & 1.4 & extractive answer \\
mt & 3500 & 500 & 1000 & 11.5 & Bn$\rightarrow$En translation \\
pp & 3500 & 500 & 1000 & 11.8 & paraphrase \\
cc & 402 & 57 & 115 & 10.1 & conversational reply \\
bt & 2642 & 378 & 755 & 1.0 & 3-way classification \\
summ & 4000 & 500 & 500 & 73.1 & news summarisation \\
\bottomrule
\end{tabular}

%% file: tables/tab_teachers.tex
\begin{tabular}{@{}lrrrrrrrrr@{}}
\toprule
& \multicolumn{3}{c}{BanglaT5} & \multicolumn{3}{c}{gemma} & \multicolumn{3}{c}{Qwen} \\
\cmidrule(lr){2-4}\cmidrule(lr){5-7}\cmidrule(lr){8-10}
task & T & CE & KD & T & CE & KD & T & CE & KD \\
\midrule
qa & 0.970 & 0.953 & 0.949 & 0.981 & 0.953 & 0.953 & 0.934 & 0.918 & 0.902 \\
mt & 0.497 & 0.394 & 0.354 & 0.552 & 0.516 & 0.514 & 0.535 & 0.437 & 0.438 \\
pp & 0.543 & 0.472 & 0.468 & 0.539 & 0.514 & 0.520 & 0.500 & 0.443 & 0.447 \\
cc & 0.636 & 0.479 & 0.471 & 0.708$^{\ast}$ & 0.717 & 0.728 & 0.723 & 0.672 & 0.684 \\
bt & 0.955 & 0.950 & 0.951 & 0.946$^{\ast}$ & 0.962 & 0.956 & 0.963 & 0.947 & 0.959 \\
\bottomrule
\end{tabular}

%% file: tables/tab_parity.tex
\begin{tabular}{@{}lrrrrr@{}}
\toprule
student & $n$ & sig. & gains & equiv.\ $\pm$.032 & equiv.\ $\pm$.020 \\
\midrule
LLM students (gemma, Qwen) & 30 & 1 & 0 & 28 & 26 \\
BanglaT5 student & 15 & 0 & 0 & 14 & 13 \\
\midrule
total & 45 & 1 & 0 & 42 & 39 \\
\bottomrule
\end{tabular}

%% file: tables/tab_cells45.tex
\begin{tabular}{@{}lllrrrrc@{}}
\toprule
student & task & gate & $\Delta$F1 & 95\% CI & $p_{\text{perm}}$ & $p_{\text{Holm}}$ & equiv. \\
\midrule
BanglaT5 & qa & entropy gate & $-0.0005$ & $[-0.0050, +0.0040]$ & 0.841 & 1.000 & \checkmark \\
BanglaT5 & qa & semantic gate & $+0.0053$ & $[+0.0009, +0.0106]$ & 0.037 & 0.552 & \checkmark \\
BanglaT5 & qa & CHAD & $+0.0031$ & $[-0.0001, +0.0072]$ & 0.104 & 1.000 & \checkmark \\
BanglaT5 & mt & entropy gate & $-0.0028$ & $[-0.0103, +0.0046]$ & 0.470 & 1.000 & \checkmark \\
BanglaT5 & mt & semantic gate & $-0.0017$ & $[-0.0118, +0.0088]$ & 0.742 & 1.000 & \checkmark \\
BanglaT5 & mt & CHAD & $-0.0018$ & $[-0.0087, +0.0050]$ & 0.605 & 1.000 & \checkmark \\
BanglaT5 & pp & entropy gate & $+0.0012$ & $[-0.0025, +0.0049]$ & 0.542 & 1.000 & \checkmark \\
BanglaT5 & pp & semantic gate & $-0.0029$ & $[-0.0072, +0.0012]$ & 0.176 & 1.000 & \checkmark \\
BanglaT5 & pp & CHAD & $-0.0016$ & $[-0.0046, +0.0017]$ & 0.331 & 1.000 & \checkmark \\
BanglaT5 & cc ($n{=}115$) & entropy gate & $-0.0045$ & $[-0.0229, +0.0130]$ & 0.624 & 1.000 & \checkmark \\
BanglaT5 & cc ($n{=}115$) & semantic gate & $-0.0158$ & $[-0.0332, +0.0015]$ & 0.078 & 1.000 & \checkmark \\
BanglaT5 & cc ($n{=}115$) & CHAD & $-0.0166$ & $[-0.0378, +0.0043]$ & 0.130 & 1.000 &  \\
BanglaT5 & bt & entropy gate & $+0.0000$ & $[-0.0053, +0.0053]$ & 1.000 & 1.000 & \checkmark \\
BanglaT5 & bt & semantic gate & $-0.0040$ & $[-0.0106, +0.0026]$ & 0.456 & 1.000 & \checkmark \\
BanglaT5 & bt & CHAD & $-0.0066$ & $[-0.0132, -0.0013]$ & 0.066 & 0.931 & \checkmark \\
\midrule
gemma & qa & EWAD-conf & $+0.0045$ & $[-0.0044, +0.0135]$ & 0.341 & 1.000 & \checkmark \\
gemma & qa & +CPDP & $-0.0000$ & $[-0.0084, +0.0085]$ & 1.000 & 1.000 & \checkmark \\
gemma & qa & CHAD & $+0.0029$ & $[-0.0048, +0.0109]$ & 0.479 & 1.000 & \checkmark \\
gemma & mt & EWAD-conf & $-0.0031$ & $[-0.0127, +0.0066]$ & 0.516 & 1.000 & \checkmark \\
gemma & mt & +CPDP & $+0.0041$ & $[-0.0054, +0.0138]$ & 0.399 & 1.000 & \checkmark \\
gemma & mt & CHAD & $-0.0102$ & $[-0.0196, -0.0007]$ & 0.034 & 0.887 & \checkmark \\
gemma & pp & EWAD-conf & $-0.0048$ & $[-0.0126, +0.0029]$ & 0.229 & 1.000 & \checkmark \\
gemma & pp & +CPDP & $-0.0047$ & $[-0.0119, +0.0025]$ & 0.202 & 1.000 & \checkmark \\
gemma & pp & CHAD & $-0.0132$ & $[-0.0205, -0.0058]$ & $<.001$ & 0.024 & \checkmark \\
gemma & cc ($n{=}115$) & EWAD-conf & $+0.0049$ & $[-0.0081, +0.0181]$ & 0.480 & 1.000 & \checkmark \\
gemma & cc ($n{=}115$) & +CPDP & $+0.0017$ & $[-0.0130, +0.0164]$ & 0.831 & 1.000 & \checkmark \\
gemma & cc ($n{=}115$) & CHAD & $-0.0025$ & $[-0.0195, +0.0146]$ & 0.792 & 1.000 & \checkmark \\
gemma & bt & EWAD-conf & $-0.0093$ & $[-0.0185, -0.0013]$ & 0.065 & 1.000 & \checkmark \\
gemma & bt & +CPDP & $-0.0106$ & $[-0.0185, -0.0040]$ & 0.008 & 0.216 & \checkmark \\
gemma & bt & CHAD & $+0.0026$ & $[-0.0040, +0.0093]$ & 0.685 & 1.000 & \checkmark \\
\midrule
Qwen & qa & EWAD-conf & $-0.0035$ & $[-0.0171, +0.0097]$ & 0.616 & 1.000 & \checkmark \\
Qwen & qa & +CPDP & $-0.0025$ & $[-0.0155, +0.0099]$ & 0.699 & 1.000 & \checkmark \\
Qwen & qa & CHAD & $+0.0192$ & $[+0.0069, +0.0320]$ & 0.003 & 0.093 & \checkmark \\
Qwen & mt & EWAD-conf & $+0.0020$ & $[-0.0074, +0.0112]$ & 0.667 & 1.000 & \checkmark \\
Qwen & mt & +CPDP & $+0.0008$ & $[-0.0085, +0.0099]$ & 0.871 & 1.000 & \checkmark \\
Qwen & mt & CHAD & $-0.0022$ & $[-0.0113, +0.0068]$ & 0.633 & 1.000 & \checkmark \\
Qwen & pp & EWAD-conf & $-0.0025$ & $[-0.0107, +0.0058]$ & 0.560 & 1.000 & \checkmark \\
Qwen & pp & +CPDP & $-0.0011$ & $[-0.0093, +0.0067]$ & 0.778 & 1.000 & \checkmark \\
Qwen & pp & CHAD & $-0.0023$ & $[-0.0090, +0.0046]$ & 0.498 & 1.000 & \checkmark \\
Qwen & cc ($n{=}115$) & EWAD-conf & $-0.0223$ & $[-0.0446, +0.0002]$ & 0.052 & 1.000 &  \\
Qwen & cc ($n{=}115$) & +CPDP & $-0.0139$ & $[-0.0313, +0.0035]$ & 0.116 & 1.000 & \checkmark \\
Qwen & cc ($n{=}115$) & CHAD & $-0.0201$ & $[-0.0389, -0.0019]$ & 0.032 & 0.867 &  \\
Qwen & bt & EWAD-conf & $-0.0040$ & $[-0.0132, +0.0040]$ & 0.545 & 1.000 & \checkmark \\
Qwen & bt & +CPDP & $+0.0013$ & $[-0.0093, +0.0119]$ & 1.000 & 1.000 & \checkmark \\
Qwen & bt & CHAD & $+0.0013$ & $[-0.0079, +0.0106]$ & 1.000 & 1.000 & \checkmark \\
\bottomrule
\end{tabular}

%% file: tables/tab_ewadfw.tex
\begin{tabular}{@{}llrrrrr@{}}
\toprule
task & model & $\Delta$ vs own CE & 95\% CI & $p_{\text{Holm}}$ & $\Delta$ vs other CE & 95\% CI \\
\midrule
qa & EWAD-conf & $-0.0058$ & $[-0.0133, +0.0013]$ & 0.625 & $+0.0068$ & $[+0.0005, +0.0138]$ \\
qa & +CPDP-geo & $-0.0038$ & $[-0.0119, +0.0041]$ & 1.000 & $+0.0088$ & $[+0.0010, +0.0172]$ \\
mt & EWAD-conf & $-0.0127$ & $[-0.0250, -0.0004]$ & 0.248 & $-0.0447$ & $[-0.0562, -0.0333]$ \\
mt & +CPDP-geo & $+0.0484$ & $[+0.0366, +0.0606]$ & $<.001$ & $+0.0164$ & $[+0.0069, +0.0261]$ \\
pp & EWAD-conf & $+0.0079$ & $[+0.0022, +0.0138]$ & 0.060 & $+0.0139$ & $[+0.0080, +0.0198]$ \\
pp & +CPDP-geo & $-0.0082$ & $[-0.0135, -0.0029]$ & 0.018 & $-0.0022$ & $[-0.0079, +0.0035]$ \\
cc & EWAD-conf & $+0.0219$ & $[-0.0062, +0.0501]$ & 0.625 & $-0.0024$ & $[-0.0305, +0.0264]$ \\
cc & +CPDP-geo & $+0.0396$ & $[+0.0151, +0.0645]$ & 0.017 & $+0.0152$ & $[-0.0131, +0.0435]$ \\
bt & EWAD-conf & $-0.0013$ & $[-0.0093, +0.0066]$ & 1.000 & $+0.0026$ & $[-0.0053, +0.0106]$ \\
bt & +CPDP-geo & $+0.0013$ & $[-0.0053, +0.0079]$ & 1.000 & $+0.0053$ & $[-0.0013, +0.0132]$ \\
\bottomrule
\end{tabular}

%% file: tables/tab_perseed_A.tex
\begin{tabular}{@{}llrrrrrrr@{}}
\toprule
& & \multicolumn{3}{c}{F1 at early stop} & \multicolumn{3}{c}{F1 at epoch 10} & \\
\cmidrule(lr){3-5}\cmidrule(lr){6-8}
task & method & s42 & s7 & s13 & s42 & s7 & s13 & ES epochs \\
\midrule
mt & CE & 0.5161 & 0.5236 & 0.5251 & 0.4808 & 0.4722 & 0.4791 & 0,0,0 \\
mt & plain KD & 0.5124 & 0.5110 & 0.5106 & 0.4987 & 0.5012 & 0.5039 & 1,0,0 \\
mt & EWAD-conf & 0.5088 & 0.5272 & 0.5280 & 0.4967 & 0.5036 & 0.4984 & 0,0,0 \\
mt & EWAD-conf+CPDP-d & 0.5145 & 0.5274 & 0.5241 & 0.4962 & 0.5007 & 0.4938 & 0,0,0 \\
mt & CHAD & 0.5172 & 0.5097 & 0.5137 & 0.5004 & 0.5027 & 0.5016 & 1,3,1 \\
mt & CE + LS 0.05 & 0.5103 & 0.5282 & 0.5077 & 0.4996 & 0.4934 & 0.4993 & 0,0,1 \\
mt & CE + LS 0.10 & 0.5052 & 0.5222 & 0.5137 & 0.4905 & 0.5014 & 0.4990 & 0,0,0 \\
mt & CE + LS 0.20 & 0.5046 & 0.5107 & 0.5173 & 0.5034 & 0.5046 & 0.4978 & 0,0,0 \\
\midrule
pp & CE & 0.5134 & 0.5139 & 0.5167 & 0.4817 & 0.4834 & 0.4877 & 0,0,0 \\
pp & plain KD & 0.5132 & 0.5066 & 0.5106 & 0.5062 & 0.5033 & 0.5070 & 0,4,4 \\
pp & EWAD-conf & 0.5184 & 0.5113 & 0.5187 & 0.5068 & 0.5101 & 0.5087 & 1,0,0 \\
pp & CHAD & 0.5097 & 0.5136 & 0.5120 & 0.5069 & 0.5004 & 0.5024 & 0,2,5 \\
\midrule
bt & CE & 0.9589 & 0.9576 & 0.9510 & 0.9523 & 0.9576 & 0.9576 & 1,4,3 \\
bt & plain KD & 0.9430 & 0.9444 & 0.9563 & 0.9550 & 0.9563 & 0.9576 & 1,0,2 \\
bt & EWAD-conf & 0.9603 & 0.9404 & 0.9497 & 0.9550 & 0.9523 & 0.9589 & 1,2,3 \\
bt & CHAD & 0.9457 & 0.9510 & 0.9523 & 0.9563 & 0.9563 & 0.9589 & 4,1,6 \\
\midrule
cc & CE & 0.7303 & 0.7179 & 0.7050 & 0.7200 & 0.7178 & 0.7174 & 6,5,5 \\
cc & plain KD & 0.7173 & 0.7179 & 0.6993 & 0.7226 & 0.7235 & 0.7207 & 5,2,0 \\
cc & EWAD-conf & 0.7216 & 0.7303 & 0.7203 & 0.7295 & 0.7193 & 0.7270 & 5,2,1 \\
cc & CHAD & 0.7168 & 0.7189 & 0.7225 & 0.7130 & 0.7171 & 0.7172 & 5,3,4 \\
\bottomrule
\end{tabular}

%% file: tables/tab_perseed_C.tex
\begin{tabular}{@{}lrrrrrr@{}}
\toprule
& \multicolumn{3}{c}{ROUGE-L} & \multicolumn{3}{c}{BERTScore} \\
\cmidrule(lr){2-4}\cmidrule(lr){5-7}
method & s42 & s7 & s13 & s42 & s7 & s13 \\
\midrule
CE & 0.2777 & 0.2814 & 0.2771 & 0.8931 & 0.8939 & 0.8930 \\
EWAD-agree & 0.2751 & 0.2731 & 0.2711 & 0.8920 & 0.8925 & 0.8923 \\
agreement-harm gate & 0.2681 & 0.2709 & 0.2647 & 0.8921 & 0.8924 & 0.8911 \\
EWAD-agree+CPDP-kl & 0.2634 & 0.2676 & 0.2553 & 0.8931 & 0.8932 & 0.8924 \\
plain KD & 0.2595 & 0.2576 & 0.2635 & 0.8919 & 0.8917 & 0.8924 \\
EWAD-agree, constant gate & 0.2423 & 0.2488 & 0.2463 & 0.8848 & 0.8879 & 0.8885 \\
\bottomrule
\end{tabular}

%% file: tables/tab_fixed_vs.tex
\begin{tabular}{@{}llrrrr@{}}
\toprule
task & method & final $\Delta$ vs CE & $p_{\text{Holm}}$ & degr.\ $\Delta$ vs plain KD & $p_{\text{Holm}}$ \\
\midrule
mt & plain KD & $+0.0239$ & 0.019 & -- & -- \\
mt & EWAD-conf & $+0.0222$ & 0.019 & $-0.0116$ & 0.560 \\
mt & EWAD-conf+CPDP-d & $+0.0195$ & 0.019 & $-0.0150$ & 0.192 \\
mt & CHAD & $+0.0242$ & 0.019 & $-0.0019$ & 1.000 \\
mt & CE & -- & -- & $-0.0342$ & 0.067 \\
\midrule
pp & plain KD & $+0.0212$ & 0.004 & -- & -- \\
pp & EWAD-conf & $+0.0243$ & 0.004 & $-0.0030$ & 0.659 \\
pp & CHAD & $+0.0190$ & 0.004 & $-0.0039$ & 0.659 \\
pp & CE & -- & -- & $-0.0257$ & $<.001$ \\
\bottomrule
\end{tabular}

%% file: tables/tab_ls.tex
\begin{tabular}{@{}lrrrrr@{}}
\toprule
method & degradation & $f$ per seed & $f$ on means & early-stop $\Delta$ vs CE & $p$ \\
\midrule
CE & $-0.0442$ $\pm$ 0.0082 & -- & -- & -- & -- \\
CE + LS 0.05 & $-0.0180$ $\pm$ 0.0146 & 1.14, 0.40, 0.96 & 0.77 & $-0.0062$ & 0.449 \\
CE + LS 0.10 & $-0.0168$ $\pm$ 0.0035 & 0.95, 0.74, 0.80 & 0.80 & $-0.0079$ & 0.252 \\
CE + LS 0.20 & $-0.0089$ $\pm$ 0.0095 & 1.58, 1.09, 0.68 & 1.03 & $-0.0107$ & 0.084 \\
plain KD & $-0.0101$ $\pm$ 0.0035 & -- & -- & -- & -- \\
\bottomrule
\end{tabular}

%% file: tables/tab_gatequality.tex
\begin{tabular}{@{}lrrrrr@{}}
\toprule
& qa & mt & pp & cc & bt \\
\midrule
helpful share & .93 & .66 & .21 & .73 & .71 \\
held-out $R^2$ & $-$.12 & .19 & .03 & $-$.24 & .00 \\
held-out AUC & .47 & .65 & .64 & .59 & .63 \\
\midrule
gemma: rank corr.\ across seeds & -- & 0.52--0.70 & 0.74--0.81 & 0.50--0.65 & 0.69--0.78 \\
\bottomrule
\end{tabular}

%% file: tables/tab_weak.tex
\begin{tabular}{@{}lrrrr@{}}
\toprule
model & $\Delta$F1 & 95\% CI & $p_{\text{Holm}}$ & $\Delta$BERTScore \\
\midrule
teacher & $-0.2909$ & $[-0.3202, -0.2624]$ & -- & $-0.0379$ \\
CE (no teacher) & $+0.0000$ & $[+0.0000, +0.0000]$ & -- & $+0.0000$ \\
plain KD & $-0.0054$ & $[-0.0134, +0.0022]$ & 0.353 & $-0.0007$ \\
entropy gate & $-0.0024$ & $[-0.0099, +0.0048]$ & 0.533 & $-0.0001$ \\
semantic gate & $-0.0094$ & $[-0.0167, -0.0027]$ & 0.034 & $-0.0014$ \\
CHAD & $-0.0064$ & $[-0.0136, +0.0002]$ & 0.220 & $-0.0010$ \\
\bottomrule
\end{tabular}

%% file: tables/tab_summ_contrasts.tex
\begin{tabular}{@{}llrrr@{}}
\toprule
metric & contrast & $\Delta$ & 95\% CI & $p_{\text{Holm}}$ \\
\midrule
rougeL & EWAD-agree, constant gate $-$ CE & $-0.0329$ & $[-0.0397, -0.0261]$ & 0.002 \\
rougeL & plain KD $-$ CE & $-0.0185$ & $[-0.0248, -0.0123]$ & 0.008 \\
rougeL & EWAD-agree+CPDP-kl $-$ CE & $-0.0166$ & $[-0.0302, -0.0031]$ & 0.095 \\
rougeL & EWAD-agree $-$ EWAD-agree, constant gate & $+0.0273$ & $[+0.0206, +0.0341]$ & 0.005 \\
rougeL & agreement-harm gate $-$ CE & $-0.0108$ & $[-0.0172, -0.0044]$ & 0.040 \\
rougeL & EWAD-agree $-$ plain KD & $+0.0129$ & $[+0.0068, +0.0191]$ & 0.027 \\
rougeL & agreement-harm gate $-$ plain KD & $+0.0077$ & $[+0.0008, +0.0146]$ & 0.095 \\
rougeL & EWAD-agree $-$ CE & $-0.0056$ & $[-0.0106, -0.0006]$ & 0.095 \\
\midrule
bertscore & EWAD-agree, constant gate $-$ CE & $-0.0063$ & $[-0.0108, -0.0018]$ & 0.178 \\
bertscore & plain KD $-$ CE & $-0.0013$ & $[-0.0023, -0.0003]$ & 0.175 \\
bertscore & EWAD-agree+CPDP-kl $-$ CE & $-0.0004$ & $[-0.0015, +0.0006]$ & 0.948 \\
bertscore & EWAD-agree $-$ EWAD-agree, constant gate & $+0.0052$ & $[+0.0004, +0.0100]$ & 0.242 \\
bertscore & agreement-harm gate $-$ CE & $-0.0015$ & $[-0.0029, -0.0000]$ & 0.242 \\
bertscore & EWAD-agree $-$ plain KD & $+0.0002$ & $[-0.0005, +0.0010]$ & 0.948 \\
bertscore & agreement-harm gate $-$ plain KD & $-0.0001$ & $[-0.0016, +0.0013]$ & 0.948 \\
bertscore & EWAD-agree $-$ CE & $-0.0011$ & $[-0.0021, -0.0001]$ & 0.242 \\
\midrule
semsim & EWAD-agree, constant gate $-$ CE & $-0.0137$ & $[-0.0170, -0.0104]$ & 0.008 \\
semsim & plain KD $-$ CE & $-0.0082$ & $[-0.0129, -0.0035]$ & 0.056 \\
semsim & EWAD-agree+CPDP-kl $-$ CE & $-0.0055$ & $[-0.0124, +0.0014]$ & 0.250 \\
semsim & EWAD-agree $-$ EWAD-agree, constant gate & $+0.0134$ & $[+0.0087, +0.0180]$ & 0.028 \\
semsim & agreement-harm gate $-$ CE & $-0.0046$ & $[-0.0087, -0.0005]$ & 0.148 \\
semsim & EWAD-agree $-$ plain KD & $+0.0078$ & $[+0.0028, +0.0129]$ & 0.061 \\
semsim & agreement-harm gate $-$ plain KD & $+0.0036$ & $[-0.0012, +0.0085]$ & 0.250 \\
semsim & EWAD-agree $-$ CE & $-0.0003$ & $[-0.0048, +0.0041]$ & 0.837 \\
\bottomrule
\end{tabular}

%% file: tables/tab_metric_agreement.tex
\begin{tabular}{@{}llrrrcc@{}}
\toprule
task & method & $\Delta$F1 & $\Delta$BERTScore & $\Delta$semsim & same sign & $|\Delta\text{F1}|<\sigma$ \\
\midrule
bt & plain KD & $-0.0079$ & $-0.0008$ & $-0.0037$ & yes & no \\
bt & EWAD-conf & $-0.0057$ & $-0.0008$ & $-0.0025$ & yes & yes \\
bt & CHAD & $-0.0062$ & $-0.0007$ & $-0.0026$ & yes & yes \\
cc & plain KD & $-0.0062$ & $-0.0010$ & $-0.0012$ & yes & yes \\
cc & EWAD-conf & $+0.0064$ & $+0.0007$ & $+0.0030$ & yes & yes \\
cc & CHAD & $+0.0017$ & $+0.0010$ & $+0.0001$ & yes & yes \\
mt & plain KD & $-0.0102$ & $-0.0012$ & $-0.0013$ & yes & no \\
mt & EWAD-conf & $-0.0003$ & $-0.0001$ & $+0.0007$ & no & yes \\
mt & CHAD & $-0.0081$ & $-0.0011$ & $-0.0011$ & yes & no \\
pp & plain KD & $-0.0045$ & $-0.0008$ & $-0.0034$ & yes & no \\
pp & EWAD-conf & $+0.0015$ & $-0.0002$ & $-0.0007$ & no & yes \\
pp & CHAD & $-0.0029$ & $-0.0007$ & $-0.0028$ & yes & yes \\
\bottomrule
\end{tabular}